\documentclass[10pt,twocolumn,letterpaper]{article}

\usepackage{cvpr}
\usepackage{times}
\usepackage{epsfig}
\usepackage{graphicx}
\usepackage{amsmath}
\usepackage{amssymb}
\usepackage{booktabs}

\usepackage[pagebackref=true,breaklinks=true,letterpaper=true,colorlinks,bookmarks=false]{hyperref}

\cvprfinalcopy 

\def\cvprPaperID{3386} 

\ifcvprfinal\fi
\begin{document}

\title{Listen to the Image\thanks{\copyright 2019 IEEE. Personal use of this material is permitted. Permission from IEEE must be obtained for all other uses, in any current or future media, including reprinting/republishing this material for advertising or promotional purposes, creating new collective works, for resale or redistribution to servers or lists, or reuse of any copyrighted component of this work in other works.}}

\author{Di Hu, Dong Wang, Xuelong Li, Feiping Nie\thanks{Corresponding author.}, Qi Wang\\
School of Computer Science and Center for OPTical IMagery Analysis and Learning (OPTIMAL),\\
Northwestern Polytechnical University\\
{\tt\small hdui831@mail.nwpu.edu.cn, nwpuwangdong@gmail.com,}\\
{\tt\small xuelong\_li@ieee.org, feipingnie@gmail.com, crabwq@gmail.com}
}

\maketitle
\thispagestyle{empty}

\begin{abstract}
Visual-to-auditory sensory substitution devices can assist the blind in sensing the visual environment by translating the visual information into a sound pattern.
To improve the translation quality, the task performances of the blind are usually employed to evaluate different encoding schemes.
In contrast to the toilsome human-based assessment, we argue that machine model can be also developed for evaluation, and more efficient.
To this end, we firstly propose two distinct cross-modal perception model w.r.t. the late-blind and congenitally-blind cases, which aim to generate concrete visual contents based on the translated sound.
To validate the functionality of proposed models, two novel optimization strategies w.r.t. the primary encoding scheme are presented.
Further, we conduct sets of human-based experiments to evaluate and compare them with the conducted machine-based assessments in the cross-modal generation task.
Their highly consistent results w.r.t. different encoding schemes indicate that using machine model to accelerate optimization evaluation and reduce experimental cost is feasible to some extent, which could dramatically promote the upgrading of encoding scheme then help the blind to improve their visual perception ability.
\end{abstract}

\section{Introduction}
There are millions of blind people all over the world, how to help them to ``re-see" the outside world is a significant but challenging task.
In general, the main causes of blindness come from various eye diseases~\cite{thylefors1995global}. That is to say, the visual cortex should be largely unwounded.
Hence, it becomes possible to use other organs (e.g., ears) as the sensor to ``visually'' perceive the environment according to the theory of cross-modal plasticity~\cite{bavelier2002cross}.
In the past decades, there have been several projects attempting to help the disabled to recover their lost senses via other sensory channels, and the relevant equipments are usually named as \emph{Sensory Substitution} (SS) devices.
In this paper, we mainly focus on the visual-to-auditory SS device of vOICe\footnote{www.seeingwithsound.com} (The upper case of OIC means ``Oh! I See!'').

\begin{figure}[t]
\centering
\includegraphics[width=8cm]{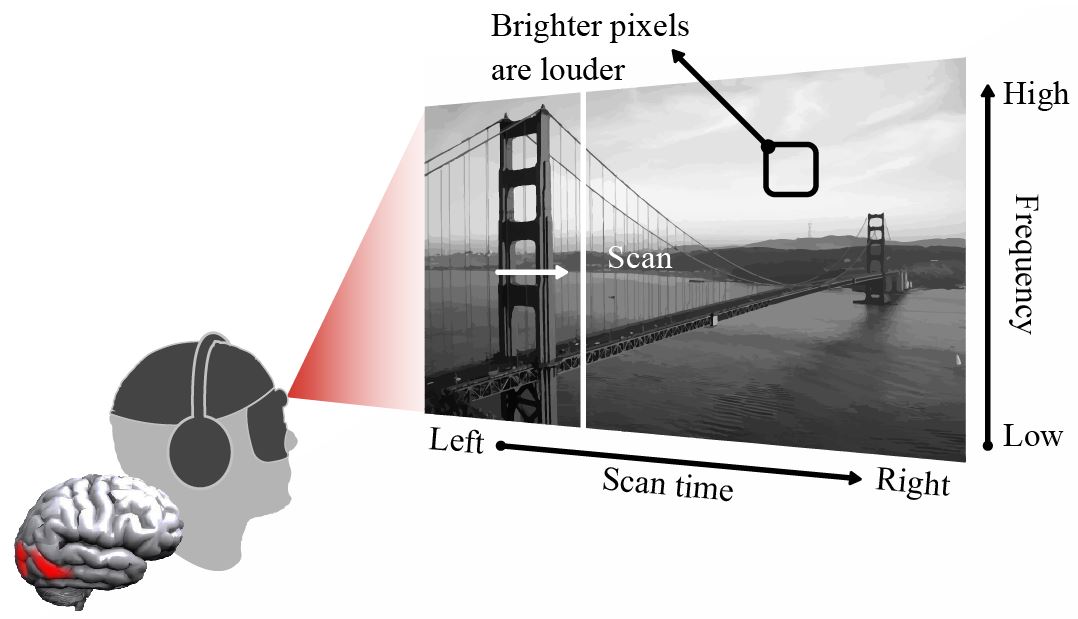}\\
\caption{An illustration of the vOICe device. A camera mounted on forehead captures the scene in front of the blind, which is then converted to sound and transmitted to headphones. After 10-15 hours of training, the regions of visual cortex become active due to cross-modal plasticity. Best viewed in color.}\label{illustrator}
\end{figure}

The vOICe is an auditory sensory substitution device that encodes 2D gray image into 1D audio signal.
Concretely, it translates vertical position to frequency, left-right position to scan time, and brightness to sound loudness, as shown in Fig.~\ref{illustrator}.
After training with the device, both blindfolded sighted and blind participants can identify different objects just via the encoded audio signal~\cite{stiles2015auditory}.
More surprisingly, neural imaging studies (e.g., fMRI and PET) show that the visual cortex of trained participants are activated when listening to the vOICe audio~\cite{amedi2007shape,poirier2007neuroimaging}. Some participants, especially the late-bind, report that they can ``see'' the shown objects, even the colorful appearance~\cite{kristjansson2016designing}.
As a consequence, the vOICe device is considered to provide a novel and effective way to assist the blind in ``visually'' sensing the world via hearing~\cite{ward2014sensory}.

However, the current vOICe encoding scheme is identified as a primary solution~\cite{stiles2015auditory}. More optimization efforts should be considered and evaluated to improve the translation quality.
Currently, we have to resort to the cognitive assessment of participants.
However, two characteristics make it difficult to be adopted in practice.
First, the human-based assessment consists of complicated training and testing procedures, during which the participants have to keep high-concentration and -pressure for a long period of time.
Such toilsome evaluation is unfriendly to the participants, and it is also inefficient.
Second, amounts of control experiments are required to provide convincing assessments, which therefore need much efforts as well as huge employment cost.
As a result, there is few works focusing on the optimization of current encoding scheme, while it indeed plays a crucial role in improving the visual perception ability of the blinds~\cite{stiles2015auditory,kristjansson2016designing}.

By contrast, with the rapid development of machine learning techniques, the artificial perception models have taken the advantages of efficiency, economical, and convenience, which exactly settle the above problems of human-based evaluation.
Hence, can we design proper machine model for cross-modal perception like human, and view it as an assessment reference for different audio encoding schemes?
In this paper, to answer these questions, we develop two cross-modal perception models w.r.t. the distinct late- and congenitally-blind case, respectively.
In the late-blind case, as they have seen the environments before they were blind. The learned visual experiences could help them imagine the corresponding objects when listening to the translated sound via the vOICe~\cite{amedi2007shape,kristjansson2016designing}.
Based on this, we propose one \emph{Late-Blind Model} (LBM), where the visual generative model is initialized from abundant visual scene for imitating the conditions before blindness, then employed for visual generation based on the audio embeddings of separated visual perception.
While the congenitally-blind have never seen the world before, therefore the absence of visual knowledge makes them difficult to imagine the shown objects.
In this case, we propose a novel \emph{Congenitally-Blind Model} (CBM), where the visual generation depends on the feedback of audio discriminator via a derivable sound translator.
Without any prior knowledge in the visual domain, the CBM can generate concrete visual contents related to the audio signal.
Finally, we employ the proposed model to evaluate different encoding schemes presented by us and compare the assessments with sets of conducted human-based evaluations.
Briefly, our contributions are summarized as follows,
\begin{itemize}
\setlength{\itemsep}{0pt}
\setlength{\parsep}{0pt}
\setlength{\parskip}{0pt}
  \item We propose a novel computer-assisted evaluation task about the encoding schemes of visual-to-auditory SS devices for the blind community, which aims to vastly improve the efficiency of evaluation.
  \item Two novel cross-modal perception models are developed for the late-blind and congenitally-blind conditions, respectively. The generated proper visual content w.r.t. the audio signal confirm their effectiveness.
  \item We present two novel optimization strategies w.r.t. the current encoding schemes, and validate them on the proposed model.
  \item We conduct amounts of human-based evaluation experiments for the presented encoding schemes.
            The highly consistent results with the machine-model verify the feasibility of machine-based assessments.
\end{itemize}

\section{Related Works}

\subsection{Sensory Substitution}
 To remedy the disabled visual perception ability of the blind, many attempts have been made to covert the visual information into other sensory perception, where the visual-to-auditory sensory substitution is the most attractive and promising fashion.
 The early echolocation approaches provide a considerable conversion manner~\cite{ifukube1991blind,kolarik2014sensory}, but the required sound emission makes it impractical in many scenarios, such as the noisy environment.
 Another important attempt is to convert the visual information into speech, e.g., the Seeing-AI project\footnote{https://www.microsoft.com/en-us/seeing-ai}.
 However, such direct semantic description is just developed for the low vision community, as it is hard to imagine the concrete visual shape for the blind people, especially the congenitally blind ones.
 And it also suffers from limited object and scene categories of the training set.
 Hence, more researchers suggest to encode the visual appearance into sequential audio signal according to specific rules, such as the vOICe.

Most researchers consider that cross-modal plasticity is the essential reason to the success of the SS devices~\cite{amedi2007shape,merabet2009functional,brown2011seeing,striem2012reading,ptito2005cross}, which is also the neurophysiological criterion for judging the effectiveness of these devices~\cite{poirier2007neuroimaging}.
Cross-modal plasticity refers that the sensory deprivation of one modality can have an impact on the cortex development of remaining modalities~\cite{bavelier2002cross}. For example, the auditory cortex of deaf subjects is activated by visual stimulate~\cite{cohen1997functional}, while the visual cortex is activated by auditory messages for the blind people~\cite{striem2012reading}.
To explore the practical effects, Arno~\emph{et al.}~\cite{arno2001occipital} proposed to detect the neural activations via \emph{Positron Emission Tomography} (PET) when the participants are required to recognize patterns and shapes with the visual-to-auditory SS devices. For both blindfolded sighted and congenitally blind participants, they found  a variety of cortex regions related to visual and spatial imagery were activated relative to the baselines of audition.
More related works can be found in~\cite{amedi2007shape,merabet2009functional}.
These evidences show that regions of the brain normally dedicated to vision can be used to process visual information passed by sound~\cite{ward2014sensory,noppeney2007effects}.
In addition, the above studies also show that there exit effective interactions among modalities in the brain~\cite{bavelier2002cross,holmes2005multisensory}.

In view that cross-modal plasticity provides a convincing explanation for sensory substitution, how to effectively utilize it for the disabled people in practice becomes the key problem.
 The visual-to-auditory SS device of vOICe has provided one feasible encoding scheme that has been verified in the practical usage.
 But it is difficult to guarantee that whether the current scheme is the most effective one.
 By conducting sets of control experiments to the current scheme, such as reversing the encoding direction (e.g., the highest pitch is set to the bottom instead of the top of image), Stiles~\emph{et al.}~\cite{stiles2015auditory} found that the primary scheme did not hold the best performance in the task of matching images and sounds, which indicates that better encoding scheme is expected to improve the cross-modal perception ability of the blind people further.
 However, the evaluation of encoding schemes based on human feedback is relatively complicated and inefficient.
 In this paper, a kind of machine model is proposed to accelerate the assessment progress, which is also convenient and economical.

\begin{figure*}[t]
\centering
\includegraphics[width=16cm]{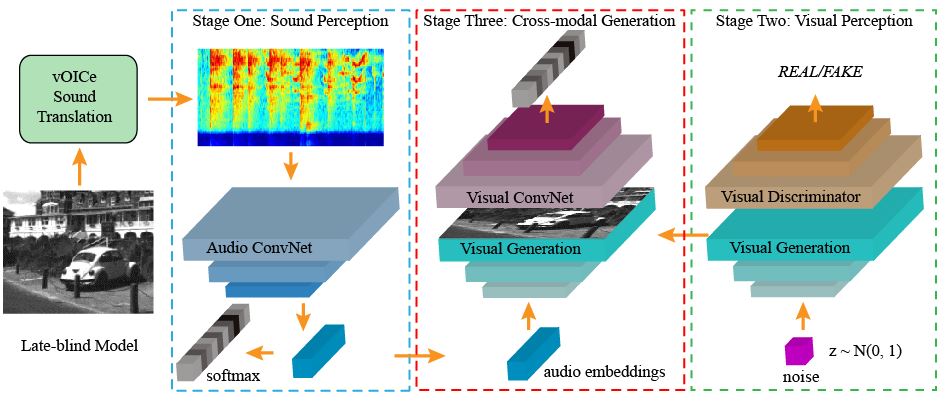}\\
\caption{The diagram of the proposed late-blind model. The image and vOICe translator outside the dashed boxes indicate the circumstance of blindness when using SS devices, while the three-stage perception model within the boxes is constituted by sound embedding, visual knowledge learning, and cross-modal generation. Best viewed in color.}\label{lbm}
\end{figure*}

\subsection{Cross-modal Machine Learning}
In the machine learning community, it is also expected to build effective cross-modal learning model similar to the brain.
And many attempts have been made in different cross-modal tasks.
Ngiam~\emph{et al.}~\cite{ngiam2011multimodal} introduced a novel learning setting to effectively evaluate the shared representation across modalities, named as ``hearing to see''.
Thereinto, the model was trained with one provided modality but tested on the other modality, which later confirmed the existing semantic correlation across modalities.
Inspired by this, Srivastava~\emph{et al.}~\cite{srivastava2012multimodal} proposed to generate the missing text for a given image, where the shared semantic provided the probability to predict corresponding descriptions.
Further, Owens~\emph{et al.}~\cite{owens2016visually} focused on the more complicated sound generation task. They presented an algorithm to synthesize sound for silent videos about hitting objects.
To make the generated sound realistic enough, they resorted to a feature-based exemplar retrieval strategy instead of direct generation~\cite{owens2016visually}.
While more recent work proposed an immediate way in generating natural sound for wild videos~\cite{zhou2017visual}.
On the contrary, Chung~\emph{et al.}~\cite{chung2017you} proposed to generate visually talking face based on an initial frame and speech signal.
For a given audio sequence, the proposed model could generate a determined image frame that best represents the speech sample at each time step~\cite{chung2017you}.
Recently, the impressive \emph{Generation Adversarial Networks} (GANs) \cite{goodfellow2014generative} provide more possibilities in the cross-modal generation.
The early works focused on the text-based image generation \cite{reed2016generative,zhang2017stackgan}, which were all developed based on the conditional GAN~\cite{mirza2014conditional}.
Moreover, it becomes possible to generate images that did not exist before.
Chen~\emph{et al.}~\cite{chen2017deep} extended above model into more complicated audiovisual perception task.
The proposed model~\cite{chen2017deep} showed noticeable performance in generating musical sounds based on the input image and vice versa.

Although the cross-modal machine learning models above have shown impressive generation ability for the missing modality, they do not satisfy the blind condition.
This is because the blind people cannot receive the visual information, while the above models have to utilize these information during training.
By contrast, we focus on this intractable cross-modal generation problem, where the missing modalities are unavailable in the whole lifecycle.

\section{Cross-modal Machine Perception}

\subsection{Late-blind Model}
People who had visual experience of outside world but go blind because of diseases or physical injury are called late-blind people.
Hence, when wearing the SS device of the vOICe to perceive objects, the pre-existing visual experience in their brain can provide effective reference for cross-modal perception.
The relevant cognitive experiments also confirm this, especially some blind participants could unconsciously color the objects while the color information was not encoded into sound~\cite{kristjansson2016designing}. This probably comes from the participants' memory of object color~\cite{kristjansson2016designing}.
Such significant character inspires us to build one three-stage \emph{Late-Blind Model} (LBM), as shown in Fig.~\ref{lbm}.
Concretely, we propose to model the late-blind case by decoupling the cross-modal perception into separated sound perception and out-of-sample visual perception, then coupling them for visual generation.
In the first stage, as the category labels of translated sound are available, the audio convolutional network of VGGish~\cite{hershey2017cnn} is employed as the perception model to learn effective audio embeddings via a typical sound classification task, where the input sound are represented in log-mel spectrogram\footnote{The mel-scale depicts the characteristics of human hearing.}.
For the visual modality, to achieve diversiform visual experience, the perception model is trained to model abundant visual imageries via adversarial mechanism in the second stage, which aims to imitate the circumstances of before going blind.
Further, the learned audio embeddings are viewed as the input to visual generation, and the whole cross-modal perception model is fine-tuned by identifying the generated images with an off-the-shelf visual classifier\footnote{The visual classifier is trained with the same dataset used in the second stage but fixed when fine-tuning.}.
Obviously, the generated images w.r.t. the translated sound should be much similar to the ones used to train the visual generator, which accordingly provides the possibilities to automatically color the shown objects.

\begin{figure}[b]
\centering
\includegraphics[width=8cm]{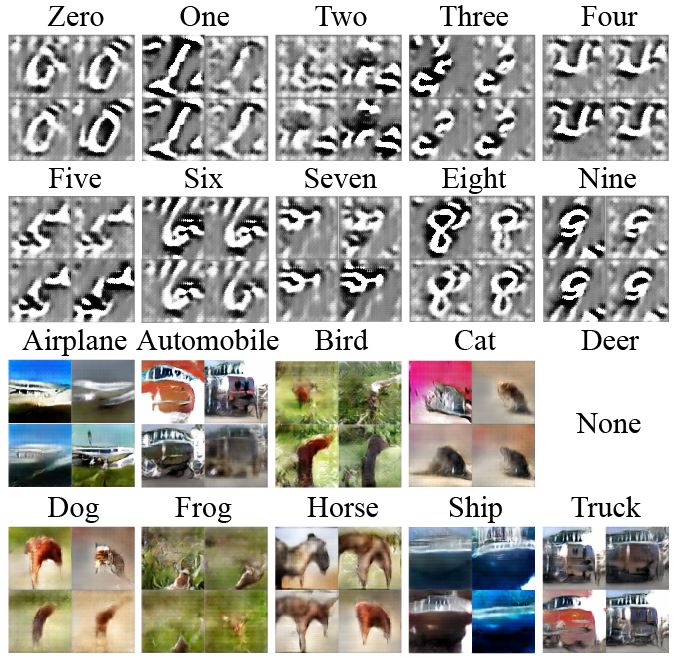}\\
\caption{Generated visual examples using our late-blind model. Top two rows are digit images generated from the translated sound of MNIST and visual knowledge of EMNIST, while bottom two rows are object images generated from the sound of CIFAR-10 and visual knowledge of ImageNet.}\label{lbmimg}
\end{figure}

\noindent{\textbf{Results.}}
To evaluate the proposed LBM, we start with the simple handwritten digits generation task.
Concretely, the MNIST images~\cite{lecun1998mnist} are firstly translated into sounds via the vOICe for training the audio subnet and the cross-modal generation model,
while the more challenging dataset of \emph{Extension MNIST} (EMNIST) Digits~\cite{cohen2017emnist} is employed for training the visual generator and classifier.
And the official training/testing splits on both datasets are adopted~\cite{lecun1998mnist,cohen2017emnist}.
Fig.~\ref{lbmimg} shows some generated digit examples w.r.t. the translated sounds in the testing set of MNIST\footnote{Network settings and training details are in the materials.}.
Obviously, the generated digit contents can be easily recognized and also corresponding to the labels of translated sounds,
which confirms that the visual experience learned from EMNIST indeed helps to build visual content with audio embeddings.
Further, we attempt to generate realistic objects by training LBM with more difficult datasets, i.e., using CIFAR-10~\cite{krizhevsky2009learning} for cross-modal generation and ImageNet \cite{deng2009imagenet} for visual knowledge learning.
To improve the complexity of visual experiences, apart from the nine categories of CIFAR-10\footnote{The deer class of CIFAR-10 is removed due to its absence in the ImageNet dataset.}, we randomly select another ten classes from ImageNet for training the visual networks.
As expected, although the translated sounds do not encode the color information of original object, the generated objects are automatically colored due to the priori visual knowledge from ImageNet, which provides a kind of confirmation to the theory of experience-driven multi-sensory associations~\cite{kristjansson2016designing}.
But on the other hand, due to the absence of real images, our generated images are not as good as the ones generated by directly discriminating images.
It exactly confirms the difficulty of cross-modal generation in the blind case.

\begin{figure*}[t]
\centering
\includegraphics[width=16cm]{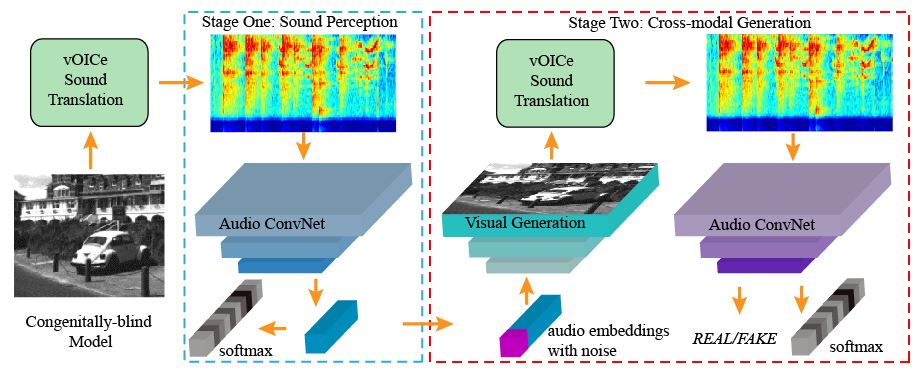}\\
\caption{The diagram of the proposed congenitally-blind model. The image and vOICe translator outside the dashed boxes represent the circumstance of blindness with SS devices, while the two-stage model within the boxes consists of preliminary sound embedding and cross-modal generative adversarial perception.}\label{cbm}
\end{figure*}

\subsection{Congenitally-blind Model}
Different from late-blind people, congenitally-blind people were born blind. Their absent visual experience makes them extremely difficult to imagine the visual appearance of shown objects.
However, cross-modal plasticity provides the possibility to effectively sense concrete visual content via the audio channel~\cite{kristjansson2016designing,ward2014sensory}, which depends on specific image-to-sound translation rules.
In practice, before training the blinds to ``see'' objects with the vOICe device, they should learn the translation rules firstly by identifying simple shapes, which could make them sense the objects more precisely~\cite{stiles2015auditory}.
In other words, the cross-modal translation helps to bridge the visual and audio perception in the brain.
Based on this, we propose a two-stage \emph{Congenitally-Blind Model} (CBM), as shown in Fig.~\ref{cbm}.
Similarly with LBM, the VGGish network is firstly utilized to model the translated sound via a classification task, and the extracted embeddings are then used as the conditional input to cross-modal generation.
In the second stage, without resorting to the prior visual knowledge, we propose to directly generate concrete visual contents by a novel cross-modal GAN, where the generator and discriminator deal with different modalities. By encoding the generated images into the sound modality with derivable cross-modal translation, it becomes feasible to directly compare the generated visual image and the original translated sound.
Meanwhile, to generate correct visual content to the translated sound, the visual generator takes the audio embeddings as the conditional input and the audio discriminator takes the softmax regression as an auxiliary classifier, which accordingly constitute a variant auxiliary classifier GAN~\cite{odena2016conditional}.

\begin{figure}[t]
\centering
\includegraphics[width=8cm]{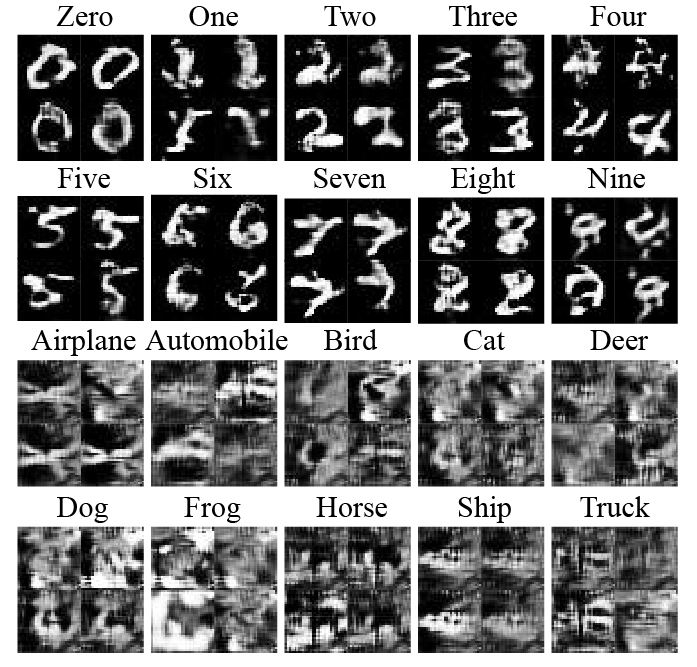}\\
\caption{Generated visual examples using our congenitally-blind model. Top two rows are digit images generated from the translated sound of the MNIST dataset, while bottom two rows are object images generated from the sound of the CIFAR-10 dataset.}\label{cbmimg}
\end{figure}

\noindent{\textbf{Results.}}
As the visual experience is not required in the congenitally-blind case, the datasets employed for pre-training the visual generator are not needed any more.
Hence, our proposed CBM is directly evaluated on MNIST and CIFAR-10 by following the traditional training/testing splits in ~\cite{lecun1998mnist, krizhevsky2009learning}.
As the vOICe translation just deals with gray images, the images discriminated by sound do not contain RGB information, as shown in Fig. \ref{cbmimg}.
Obviously, these image samples on both datasets are not as good as the ones of LBM, due to the compressed visual content by the sound translator and the absent visual experience.
Even so, our CBM can still generate concrete visual content according to the input sound, e.g, the distinct digit forms.
As the objects in CIFAR-10 suffer from complex background that distracts the sound translation for objects, the generated appearances become low-quality while the outlines can be still captured, such as horse, airplane, etc.
On the contrary, clean background can dramatically help to generate high-quality object images, and more examples can be found in the following sections.


\section{Evaluation of Encoding Schemes}  
In view that Stiles~\emph{et al.} has shown the current encoding scheme can be further optimized to improve its applicability for the blind community~\cite{stiles2015auditory},
it becomes indispensable to efficiently evaluate different schemes according to the task performance of cross-modal perception.
Traditionally, the evaluation has to be based on the inefficient participants' feedback.
In contrast, as the proposed cross-modal perception model has shown impressive visual generation ability, especially the congenitally-blind one, it is worth considering whether the machine model can be used to evaluate the encoding scheme. More importantly, such machine-based assessment is more convenient and efficient compared with the manual fashion.
Hence, in this section, we make a comparison between machine- and human-based assessment, which is performed with modified encoding schemes.

Different from the simple reversal of the encoding direction~\cite{stiles2015auditory}, we aim to explore more possibilities in optimizing the primary scheme.
First of all, a well-known fact is that there exist large differences in bandwidth between vision and audition~\cite{kristjansson2016designing}.
When 2D images are projected into 1D audio sequence with limited length, amounts of image content and details are compressed or declined.
One direct approach is to increase the audio length, which accordingly makes the vOICe encode more detailed visual contents with more audio frames.
Meanwhile, the blind people can also have more time to imagine the corresponding image content.
But on the other hand, such augmentation cannot be unlimited for efficiency.  
Hence, the time length is doubled from the primary setting of 1.05s to 2s in the first modified encoding scheme.

\begin{figure}[b]
\centering
\includegraphics[width=7.8cm]{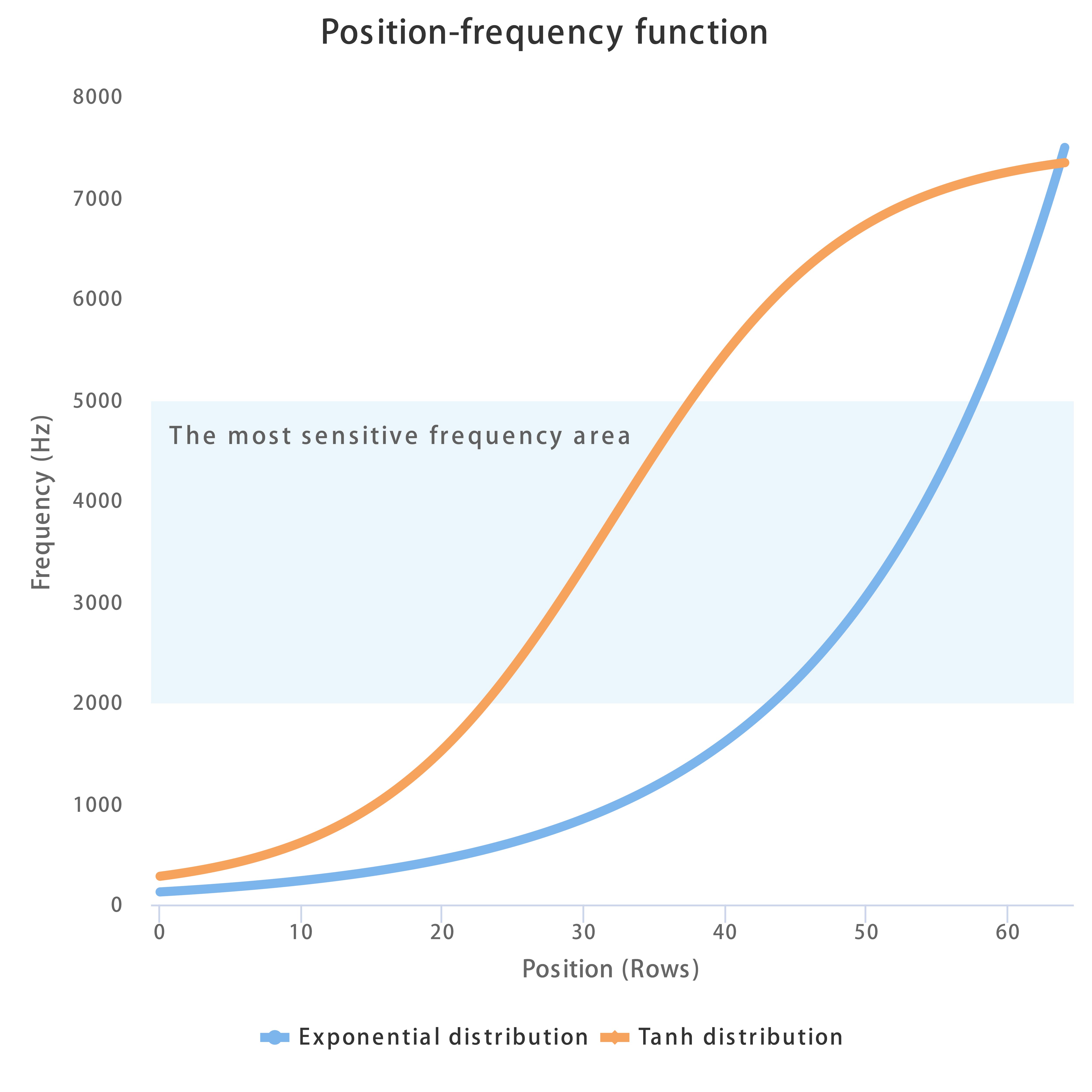}\\
\caption{Different position-frequency functions of the vOICe translation.}\label{pff}
\end{figure}

Apart from the bandwidth discrepancy, another crucial but previously neglected fact should be also paid attention to.
Generally, humans are most sensitive (i.e. able to discern at the lowest intensity) to the sound frequencies between 2K and 5K Hz~\cite{gelfand2001essentials}.
As the blind participants perceive the image content via their ears, it is necessary to provide high-quality audio signal that should exist in the sensitive frequency area.
However, due to the bandwidth discrepancy between modalities, it is difficult to precisely perceive and imagine all the visual contents in front of the blind via the translated sound.
To address this issue, we argue that the center of the visual field should be more important than other visual areas for the convenience of practical use.
Hence, we aim to effectively project the central areas to the sensitive frequencies of human ears.
Inspired by~\cite{hu2018deep}, a novel rectified tanh distribution is proposed for \emph{Position-Frequency} (PF) projection, i.e.,
\begin{equation}\label{frequency}
frequency = {s \mathord{\left/
 {\vphantom {s 2}} \right.
 \kern-\nulldelimiterspace} 2} \cdot \tanh \left( {\alpha  \cdot \left( {i - {{rows} \mathord{\left/
 {\vphantom {{Rows} 2}} \right.
 \kern-\nulldelimiterspace} 2}} \right)} \right){\rm{ + }}{s \mathord{\left/
 {\vphantom {s 2}} \right.
 \kern-\nulldelimiterspace} 2},
\end{equation}
where $s$ is the frequency range of the encoded sound, $\alpha$ is the scaling parameter, $i$ is the position of translated pixel, and $rows$ is the image height.
As shown in Fig.~\ref{pff}, it is obvious that the image centers (row 20-40) fall into the most sensitive frequencies area, where the highest pitch corresponds to the top position of image.
More importantly, compared with the suppressed frequency response of the peripheral regions of images, the central regions enjoy larger frequency range and are accordingly given more attention.
In contrast, the exponential function adopted in the primary setting takes no consideration of the auditory perception characteristics of humans.
The translated sound of most image areas are suppressed in the low-frequency area of below 2K Hz, which neither focus on the sensitive frequencies area nor highlight the central regions of images.
Hence, such function could be not an appropriate choice.

\begin{figure*}[t]
\centering
\includegraphics[width=16cm]{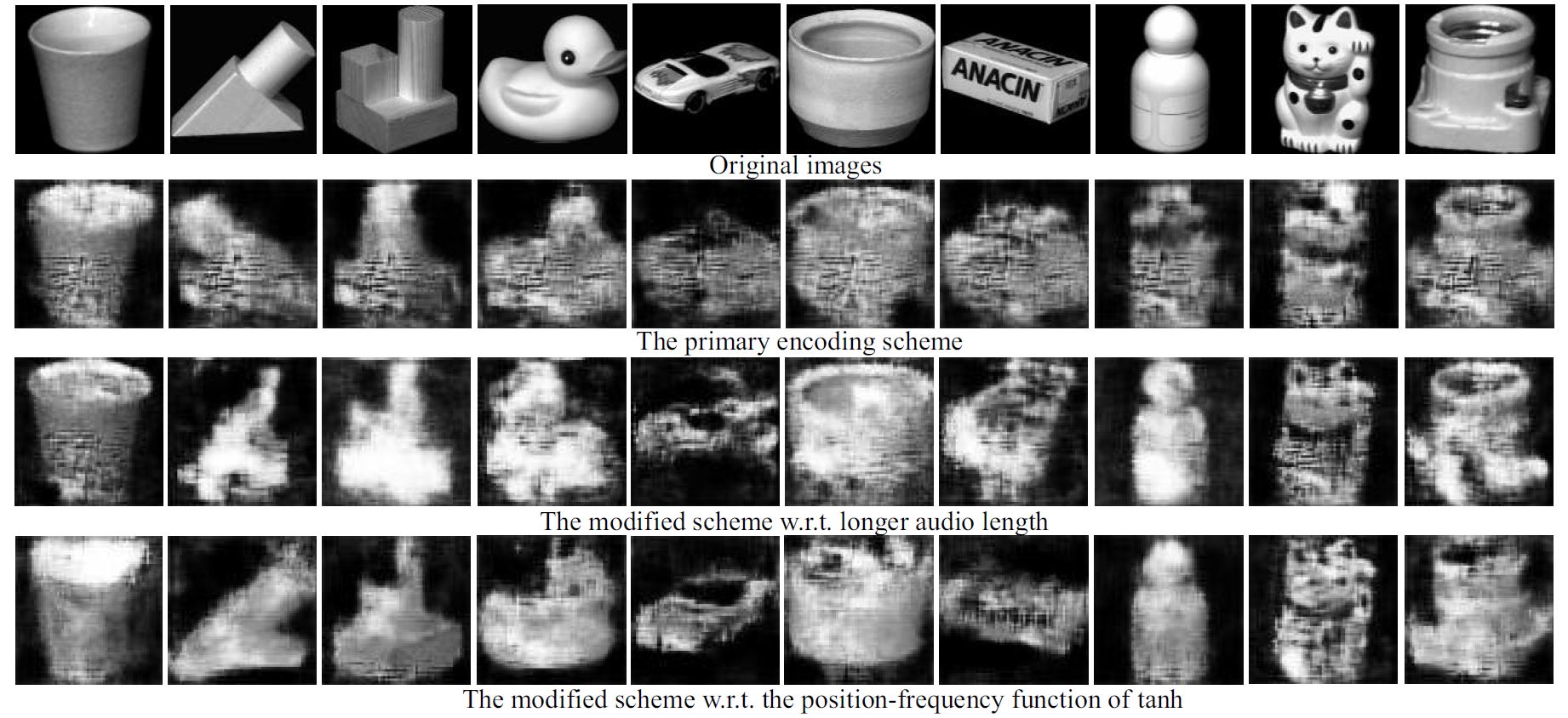}\\
\caption{Comparison among the generated image examples using our congenitally-blind model in terms of different encoding schemes.}\label{images}
\end{figure*}

To effectively evaluate the proposed encoding schemes, amounts of evaluation tests should be conducted with participants.
Accordingly, if more modified schemes are provided, much more training and testing efforts will be required, which could go beyond what we can support.
Hence, we choose to focus on these two modifications w.r.t. audio length and position-frequency function, as well as the primary one.

\subsection{Machine Assessment}
The quality of generated modality depends on the quality of the other encoded modality in the cross-modal generative model~\cite{zhou2017visual,reed2016generative,chen2017deep}.
Hence, it can be expected to evaluate different encoding schemes by comparing the generated images.
In this section, the proposed CBM is chosen as the evaluation reference, as the encoding scheme directly impacts the quality of translated sound for the audio discriminator, then further affects the performance of visual generator, as shown in Fig.~\ref{cbm}.
However, the adopted MNIST and CIFAR-10 dataset suffer from absent real object or quite complex background.
These weaknesses make it quite hard to effectively evaluate the practical effects of different encoding schemes.
Therefore, we choose the \emph{Columbia Object Image Library} (COIL-20)~\cite{nene1996columbia} as the evaluation dataset, which consists of 1,440 gray-scale images that belong to 20 objects.
These images are taken by placing the objects in the frame center in front of a clean black background, meanwhile a fixed rotation of 5 degree around the objects is also performed, which leads to 72 images per object.
Considering that this dataset will be also adopted by human-based assessment, we select 10 object categories from COIL-20 for efficiency (i.e., COIL-10), where the testing set consists of 10 selected images of each object, and the remaining ones constitute the training set.


\begin{figure}[b]
\centering
\includegraphics[width=7.8cm]{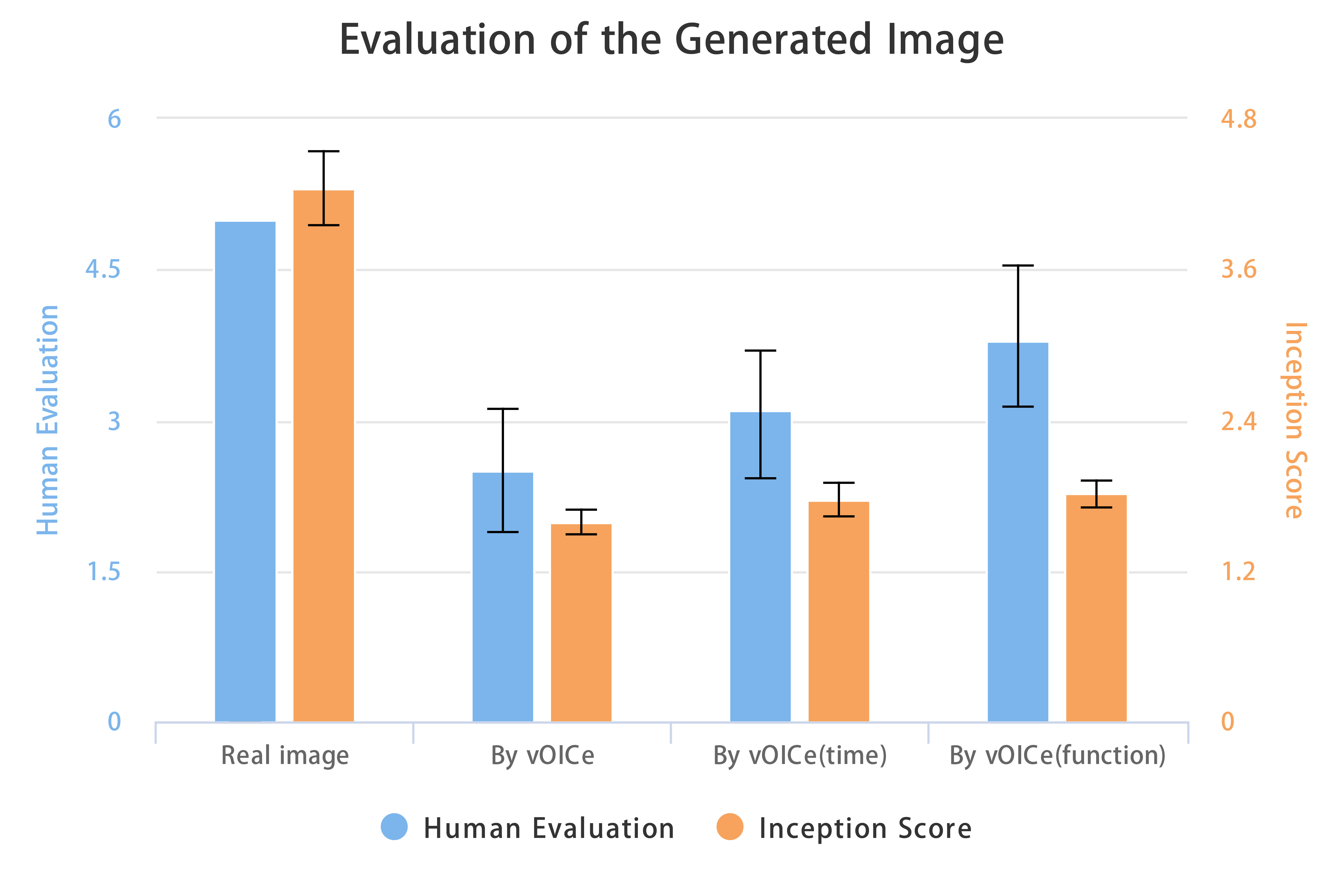}\\
\caption{Evaluation of the generated images by our CBM, where different encoding schemes are compared in human evaluation and inception score.}\label{imagebar}
\end{figure}

\noindent{\textbf{Results.}}
To comprehensively evaluate the generated images, qualitative and quantitative evaluation are both considered.
As shown in Fig.~\ref{images}, in general outlook, as well as in matters of detail, the images of the modified schemes are superior to the primary ones.
Obviously, the images generated by the primary scheme suffer from horizontal texture noise. This is because most image area are suppressed in the low-frequency domain, which makes the audio network difficult to achieve effective embeddings for visual generation. By contrast, longer audio track or more effective PF function can contribute to settle such issue.
In addition, compared with longer audio signal, the improvement attainable with the proposed PF function of tanh is more significant, such as the details of toy car and fortune cat.
And such superiority comes from the effective frequency representation of pixel positions.

For the quantitative evaluation, we choose to compute inception score~\cite{salimans2016improved} and human-based evaluation, as shown in Fig.~\ref{imagebar}.
Concretely, we ask 18 participants to grade the quality of generated images from 1 to 5 for each scheme, which correspond to \{\emph{beyond recognition, rough outline, clear outline, certain detail, legible detail}\}. Then, mean value and standard deviation are computed for comparison.
In Fig.~\ref{imagebar}, it is clear that the qualitative and quantitative evaluation show consistent results.
In particular, both of the inception score and human evaluation show that the modified encoding scheme of PF function enjoys the largest improvements, which further confirms the significance of the projection function.
And the enlarged audio length indeed helps to refine the primary scheme.

\subsection{Cognitive Assessment}
The blind participants' feedback or task performance is usually served as the indicator to the quality of encoding schemes in the conventional assessment~\cite{bach2003sensory,stiles2015auditory}.
Following the conventional evaluation strategy~\cite{stiles2015auditory}, 9 participants are randomly divided into three groups (3 participants per group). Each group corresponds to one of the three encoding schemes. The entire evaluation process takes about 11 hours.
Before the formal evaluation, the participants are firstly asked to complete the preliminary training lessons to be familiar with the translation rules and simple visual concepts.
The preliminary lessons include: identification of simple shape, i.e., triangle, square, and circle; recognition of complex shape, e.g., a normal ``L'', an upside-down ``L'', a backward ``L'', and a backward and upside-down ``L''; perception of orientation, e.g., straight white line of fixed-length in different rotation angles; estimation of lengths, e.g., horizonal white line with different lengths; localization of objects, i.e., circles in different places of images. During training, the assistant of each participant plays the pre-translated sound for them, then tell them the concrete visual content within the corresponding image\footnote{The training details and image samples can be found in material.}.
After finishing amounts of repetitive preliminary lessons, the participants are scheduled to achieve advanced training of recognizing real objects.
Concretely, the COIL-10 dataset is adopted for training and testing the participants, where the training procedure is the same as the preliminary lessons.
Note that the evaluation test is conducted after finishing the training of each object category instead of all the categories.
Finally, the evaluation results are viewed as the quality estimation of the encoded sound and used as the reference for machine-based assessment.

\begin{figure}[t]
\centering
\includegraphics[width=7.8cm]{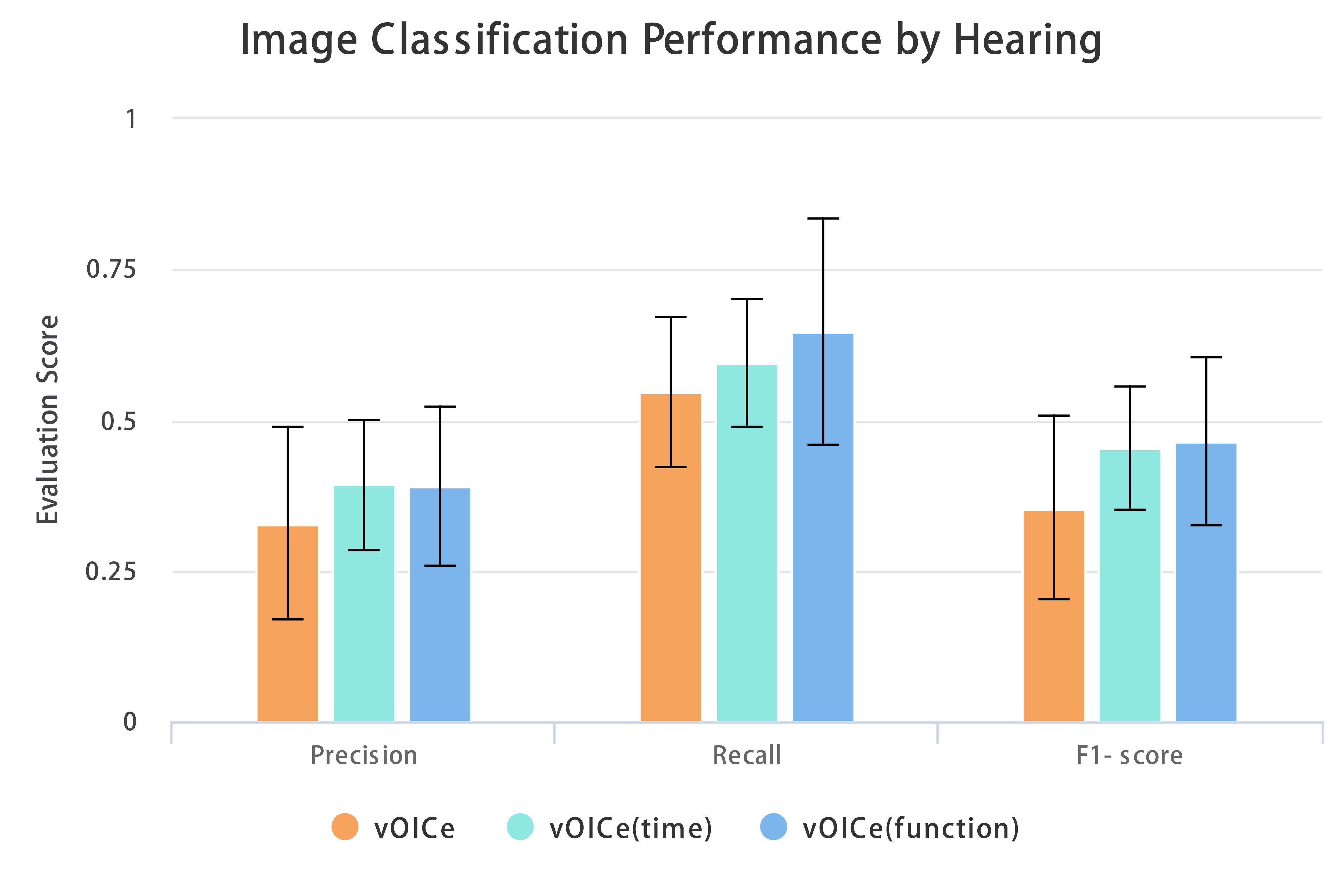}\\
\caption{Human image classification performance by hearing the translated sound via different encoding schemes.}\label{human}
\end{figure}

\noindent{\textbf{Results.}}
Fig.~\ref{human} shows the evaluation results in precision, recall and F1-score w.r.t. different encoding schemes.
Specifically, the scheme with modified PF function performed significantly better than the primary scheme ($p<0.005$, with Bonferroni multiple comparisons correction) in precision, and the recall bars indicate that the scheme with modified PF function performed significantly better than the primary scheme ($p<0.01$, with Bonferroni multiple comparisons correction). Similar results are observed on F1-score. According to the conventional criterion~\cite{stiles2015auditory}, it can be concluded that the introduced modifications indeed help to improve the quality of cross-modal translation.
And it further confirms the assumptions about audio length and the characteristics of human hearing.
Nevertheless, there still remains a large disparity of classification performance between normal visual and indirect auditory perception.
Hence, more effective encoding schemes are expected for the blind community.

\begin{table}[h]
\begin{center}
\begin{tabular}{ccc}
\toprule[2pt]
 Correlation Coefficient   & Machine (IS) & Machine (Eva.) \\
\midrule[1pt]
 Human (Recall)  & 0.947 & 1.000 \\
 Human (Precision)  & 0.952 & 0.805 \\
 Human (F1-score)  & 0.989 & 0.889 \\
\bottomrule[2pt]
\end{tabular}
\end{center}
\caption{\label{Table1} The comparison analysis between machine- and human-based assessment in terms of correlation coefficient.
}
\end{table}

\subsection{Assessment Analysis}
The main motivation of designing the cross-modal perception model is to liberate the participants from the boring and inefficient human assessments.
According to the shown results above, we can find that the modified scheme of PF function gets the best performance while the primary scheme is the worst one on both assessments.
Further, quantitative comparison is also provided in Table \ref{Table1}. 
Obviously, both assessments have reached a consensus in terms of correlation coefficient, especially the column of \emph{Inception Score} (IS), which confirms the validity of machine-based assessments to some extent.



\section{Conclusion}
In this paper, we propose a novel and effective machine-assisted evaluation approach to the visual-to-auditory SS scheme.
Compared with the conventional human-based assessments, the machine fashion performs more efficiently and conveniently.
On the other hand, this paper gives a new direction for the auto-evaluation of SS devices through limited comparisons, more possibilities should be explored in the future, including more optimization schemes and more effective machine evaluation model.
Further, the evaluation should be combined with derivable encoding module to constitute a completely automated solver of the encoding scheme without any human intervention, just like seeking high-quality models via AutoML.

\section*{Acknowledgments}
This work was supported in part by the National Natural Science Foundation of China grant under number 61772427, 61751202, U1864204 and 61773316, Natural Science Foundation of Shaanxi Province under Grant 2018KJXX-024, and Project of Special Zone for National Defense Science and Technology Innovation.

{\small
\bibliographystyle{ieee}
\bibliography{voice}
}

\clearpage

\section{Network Setting and Training}
\subsection{Late-blind Model}
In this section, we provide the architectural details of the proposed late-blind model. First, the audio ConvNet follows the VGGish architecture proposed in~\cite{hershey2017cnn}, which achieves excellent results on audio classification task. Second, the visual generator and discriminator mostly adopt the same networks in DCGAN framework~\cite{radford2015unsupervised}, except that the out\_channel size of last deconvolution layer is set to 1 in MNIST experiments. Third, due to different complexities of MNIST and CIFAR-10 data, we use different visual classifiers for those two datasets. Specifically, a small ConvNet with two convolution layers followed by two fully-connected layers is utilized to classify MNIST digits, and the ResNet-18~\cite{he2016deep} is employed as the visual classifier for CIFAR-10 dataset.

The training procedure of the whole late-blind model is composed of three steps. First, the audio ConvNet is pre-trained on MNIST/CIFAR-10 dataset for audio classification, where the audio is obtained by transforming the images with vOICe. And the visual classifier is pre-trained on EMNIST/ImageNet dataset for image classification. Second, the adversarial training strategy in~\cite{radford2015unsupervised} is used to train visual generator and discriminator on EMNIST/ImageNet dataset for visual knowledge learning. Finally, the audio ConvNet, visual generator, and visual classifier are concatenated for cross-modal generation. We firstly fix the visual generator and visual classifier, and fine-tune the audio ConvNet with image classification loss for several epochs. Then we train the visual generator and audio ConvNet together with fixed visual classifier. To be specific, we use a small initial learning rate of 0.001 with Adam optimizer~\cite{kingma2014adam} for fine-tuning  the audio ConvNet, which decreases by $\frac{1}{{10}}$ when train the visual generator and audio ConvNet together.

\subsection{Congenitally-blind Model}
The proposed congenitally-blind model consists of one sound perception module and one cross-modal generation module.
For the former, the off-the-shelf large-scale audio classification network of VGGish~\cite{hershey2017cnn} is employed, but the embedding\_dim of the second FC layer is set to 128 and the out\_dim is set to the number of classes, i.e., 10. To effectively train such sound model, we set batch\_size to 100 and choose the Adam optimizer with learning rate of 0.0002 and beta\_1 of 0.5.
For the latter, we propose a variant \emph{Auxiliary Classifier GAN} (ACGAN)~\cite{odena2016conditional}, where the input conditional label is replaced with audio embeddings.
More importantly, different from the unimodal processing of ACGAN, the generator deals with visual information while the discriminator focuses on the audio messages.
Concretely, the visual generator firstly projects and reshapes the input audio embeddings and noise into certain image shapes (e.g., $8\times8\times128$ for the MNIST dataset) via one \emph{Fully Connected} (FC) layer and one reshape layer, which is then processed by 3 up-sampling module. Each up-sampling layer is followed by one convolutional layer, as well as batch normalization and ReLU activation. The last layer projects the generated samples into single channel images (in gray scale), where sigmoid function is adopted for activation.
The audio discriminator is developed based on the VGGish network, where the activation function of convolutional layers is replaced with Leaky ReLU (with 0.2 alpha) and the discrimination layer and softmax layer are directly performed over the last Flatten layer. The entire cross-modal generation model is optimized via the Adam optimizer (learning rate is set to 0.00002 and beta\_1 is 0.5), and the batch\_size is set to 100.
Moreover, the derivable vOICe translation is derived from the official encoding scheme, and we refactor the official code into a computational graphs for the derivable purpose.

\begin{figure}[h]
\centering
\includegraphics[width=8cm]{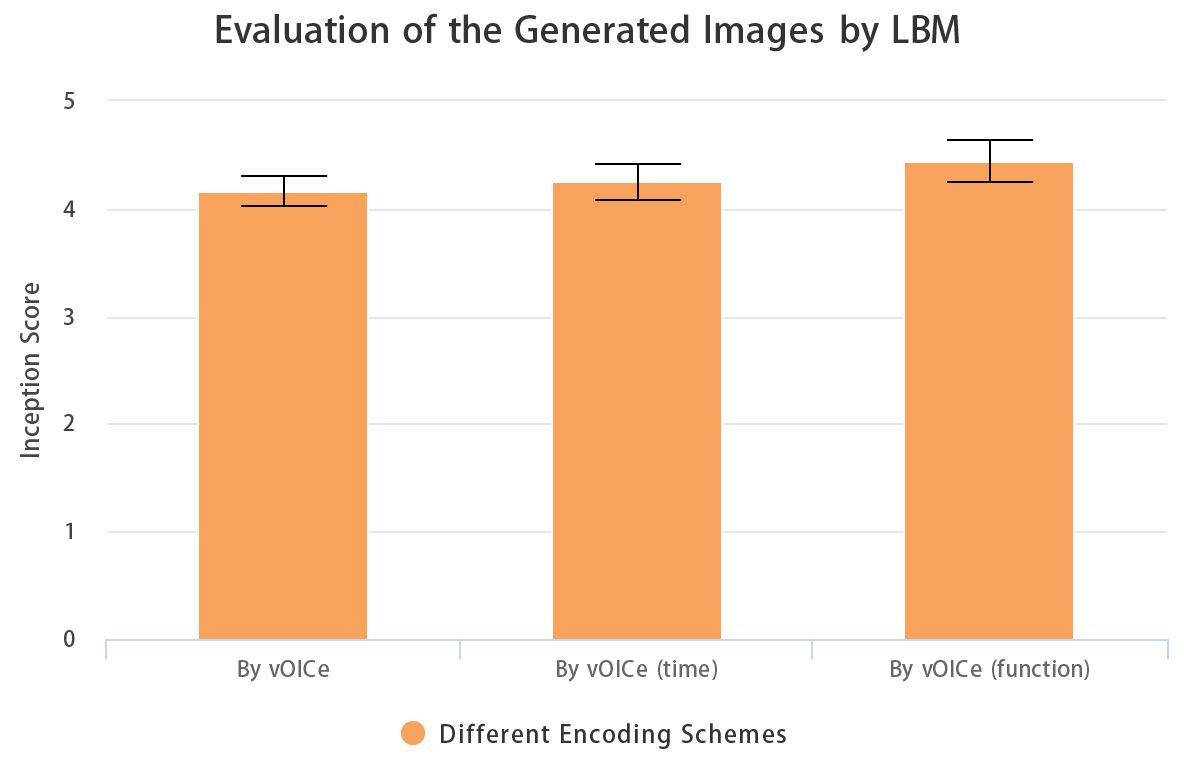}\\
\caption{Inception scores of the generated images by our LBM with different encoding schemes.}\label{incepScore}
\end{figure}

\section{Encoding Scheme Evaluation by LBM}
In this section, we evaluate different encoding schemes quantitatively and qualitatively. As shown in Fig.~\ref{incepScore}, we firstly compute inception score of intermediate generated images by our LBM with different encoding schemes. And the results show similar improvements, i.e., the modified encoding scheme of PF function achieves the largest improvements, with quantitative evaluation by CBM and human-based evaluation, which indicates that our LBM framework can also be used for machine-based assessments to some extent.
For qualitative evaluation, we show more generated image examples using our late-blind model with different encoding schemes in Fig.~\ref{images}. Generally, the images of the modified schemes are better than the primary ones in most classes on MNIST/CIFAR-10 datasets. The generated digit images using the modified encoding scheme show more clean background than the primary ones, especially in number 2, 3, 4, and 6. In addition, compared to longer audio length, the proposed PF function of tanh achieves more significant improvement in almost every class, which agrees with the quantitative results in Fig.~\ref{incepScore}. As for CIFAR-10 dataset, images of the modified schemes contain more detail information, such as windows in airplane images, legs in dog images and meadows in horse images. Meanwhile, there is no obvious improvements in several difficult classes, like automobile, ship, and truck, which confirms the difficulty of realistic objects generation. This is because the LBM learns visual generator and visual classifier from EMNIST/ImageNet datasets and it's hard to transfer learned knowledge to MNIST/CIFAR-10 datasets in cross modal generation, especially when there are extra more classes in ImageNet dataset. Moreover, as for failure case, the modified schemes obtain worse results in number 8, bird, and frog classes, and the reason behind this could be the trained LBM tend to capture detail structure of objects, resulting in overall object contour missing.

\section{Dataset Examples}
\begin{figure}[h]
\centering
\includegraphics[width=8cm]{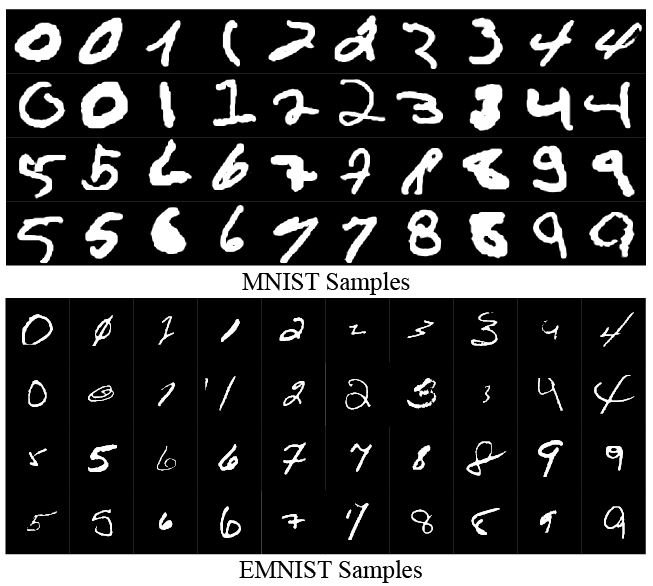}\\
\caption{Some samples in the MNIST and EMNIST Digits dataset.}\label{examples}
\end{figure}

In this section, we show some digit examples of MNIST~\cite{lecun1998mnist} and EMNIST~\cite{cohen2017emnist} in Fig.~\ref{examples}.
The MNIST dataset is a subset of a much larger dataset of NIST Special Database 19~\cite{grother1995nist}, while EMNIST is a extended MNIST dataset and a variant of the full NIST dataset.
Hence, the EMNIST Digits enjoy an increased variability (e.g., size, style, rotation, etc.) and are more challenging~\cite{cohen2017emnist}.
In the handwritten digits generation task, the EMNIST Digits are adopted to provide abundant digit samples for training the visual models of LBM, while the MNIST dataset is used for training the cross-modal perception model.

\begin{figure*}[t]
\centering
\includegraphics[width=17cm]{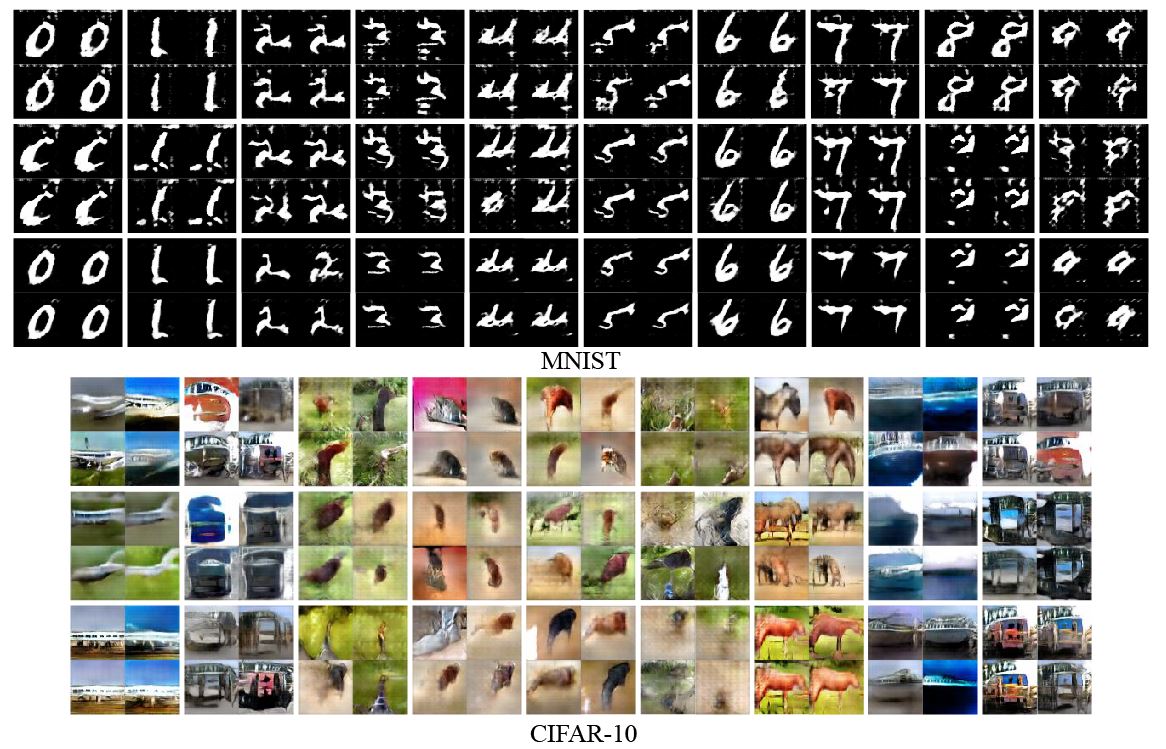}\\
\caption{Comparison among the generated image examples of MNIST/CIFAR-10 dataset using our late-blind model in terms of different encoding schemes. For each dataset, the first row represents the primary encoding scheme, the second row represents the modified scheme w.r.t. longer audio length, and third row for the modified scheme w.r.t. the position-frequency function of tanh.}\label{images}
\end{figure*}

\section{Cognitive Evaluation Details}
\subsection{The preliminary training lessons}
\begin{figure}[h]
\centering
\includegraphics[width=8cm]{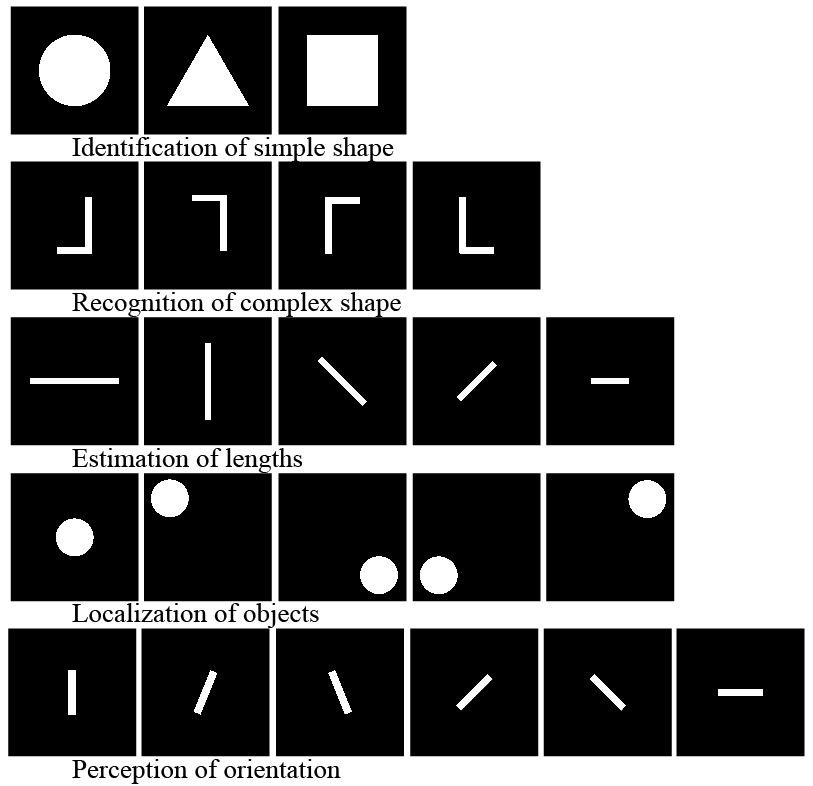}\\
\caption{The images used for training the participants in the preliminary training lessons.}\label{cog}
\end{figure}

In this section, we briefly introduce the preliminary training lessons employed for training the participants. The used images are shown in Fig.~\ref{cog}, and the corresponding sounds can be found in the ``examples'' folder.

\noindent{\textbf{The first lesson.}}
This lesson focused on the initial identification of simple shapes, i.e., circle, triangle, and square. The assistant of each participant randomly selected and played one translated sound for the participant.
After playing one sound, the participants were told the concrete content of corresponding image. During the whole first training lesson, each sound (with image) should be played for 15 times. Hence, all the translated sounds were played for 45 times totally. After finishing this lesson, the participants should take a rest for 5 minutes.

\noindent{\textbf{The second lesson.}}
Based on the first lesson, the second lesson aimed at the recognition of more complex shapes, i.e., a normal ``L'', an upside-down ``L'', a backward  ``L'', and a backward and upside-down  ``L'' (i.e., 7).
The assistants randomly selected and played translated sounds for each participant. After playing each sound, the assistants told the participants the concrete shape of corresponding image. In the second lesson, each sound should be played for 15 times totally. Hence, all the translated sounds were played for 60 times. After this lesson, the participants should take a rest for 5 minutes.

\noindent{\textbf{The third lesson.}}
In the third lesson, we aimed at the perception of orientation, i.e., straight white bar of fixed-length at 0, 22, -22, 45, -45, or 90 degrees relative to vertical (The positive angles correspond to clockwise rotations).
The assistants randomly selected and played translated sounds for each participant. After playing each sound, the participants were told the concrete orientation of corresponding bar. In the third lesson, each sound should be played for 15 times totally. Hence, all the translated sounds were played for 90 times. After this lesson, all the participants should take a rest for 5 minutes.

\noindent{\textbf{The fourth lesson.}}
This lesson focused on the estimation of different lengths, i.e., five bars with different lengths. To improve the sensitivity of lengths, these five bars were also placed in one of four orientations, i.e., 0. 90, 45, and -45 degrees as in the third lesson. During training, the assistants randomly selected and played translated sounds for each participant. After playing each sound, the assistants told the participants the concrete length of corresponding bar (by touch). The translated sound of each image should be played for 15 times totally. Hence, all the sounds were played for 75 times. After finishing this lesson, the participants should take a rest for 5 minutes.

\noindent{\textbf{The fifth lesson.}}
In the last lesson, the participants were trained to possess the localization ability, where circles in different places of images (i.e., upper-left, upper-right, bottom-left, bottom-right, and center) were considered.
During training, the assistants first randomly selected and played translated sounds for each participant. After playing each sound, the participants were told the position of corresponding circle. The translated sound of each image should be played for 15 times totally. Hence, all the sounds were played for 75 times.

\subsection{The advanced training lessons}
In the advanced training lessons, all the participants were asked to perform the image classification task by hearing the translated sounds, where the COIL-10 dataset was employed for training and testing.
Concretely, the COIL-10 dataset consisted of 10 real objects, such as toy car, fortune cat, bottle, etc.
In the training set, each category had 70 image-sound pairs. Before training the participants, they should be told that the images of each object were taken from different angles.
Note that the evaluation test was conducted after finishing the training of each object category instead of all the categories.
During training, the assistant played the translated sounds of each object for each participant in a certain order. After playing each sound, the assistant told the participant the concrete object and corresponding angle.
In the testing process, 100 sounds (of 10 objects) in the testing set were successively played for the participants, then the participants were asked to answer if the played sound corresponded to the object in the training process. After training and testing all the 10 objects, we evaluated the classification performance in terms of recall, precision, and F1-score.

\end{document}